\documentclass[letterpaper]{article} 
\usepackage{aaai2026}  
\usepackage{times}  
\usepackage{helvet}  
\usepackage{courier}  
\usepackage[hyphens]{url}  
\usepackage{graphicx} 
\usepackage{natbib}  
\usepackage{caption} 
\usepackage{algorithm}
\usepackage{algorithmic}
\usepackage{amsmath}
\usepackage{bm}
\usepackage{pifont}
\usepackage{multirow}
\usepackage{multicol}
\usepackage{amssymb}
\usepackage{array}
\usepackage{xcolor}
\usepackage{caption}
\usepackage{newfloat}
\usepackage{listings}
\DeclareCaptionStyle{ruled}{labelfont=normalfont,labelsep=colon,strut=off} 
\floatstyle{ruled}
\newfloat{listing}{tb}{lst}{}
\floatname{listing}{Listing}
\title{ProPL: Universal Semi-Supervised Ultrasound Image Segmentation via Prompt-Guided Pseudo-Labeling}
\author{
    Yaxiong Chen\textsuperscript{\rm 1}, Qicong Wang\textsuperscript{\rm 1}\thanks{Work done during an internship at MedAI Technology (Wuxi) Co. Ltd.}, Chunlei Li\textsuperscript{\rm 2}, Jingliang Hu\textsuperscript{\rm 2}, Yilei Shi\textsuperscript{\rm 2}, Shengwu Xiong\textsuperscript{\rm 1}, Xiao Xiang Zhu\textsuperscript{\rm 3}, Lichao Mou\textsuperscript{\rm 2}\thanks{Corresponding author.}
}
\affiliations{
    \textsuperscript{\rm 1} Wuhan University of Technology\\
    \textsuperscript{\rm 2} MedAI Technology (Wuxi) Co. Ltd.\\
    \textsuperscript{\rm 3} Technical University of Munich\\
    \{chenyaxiong, wqc2023302901, xiongsw\}@whut.edu.cn, \{ chunlei.li,  jingliang.hu,  yilei.shi, lichao.mou\}@medimagingai.com,  xiaoxiang.zhu@tum.de

}

\usepackage{bibentry}

\begin{document}

\maketitle

\begin{abstract}
Existing approaches for the problem of ultrasound image segmentation, whether supervised or semi-supervised, are typically specialized for specific anatomical structures or tasks, limiting their practical utility in clinical settings. In this paper, we pioneer the task of universal semi-supervised ultrasound image segmentation and propose ProPL, a framework that can handle multiple organs and segmentation tasks while leveraging both labeled and unlabeled data. At its core, ProPL employs a shared vision encoder coupled with prompt-guided dual decoders, enabling flexible task adaptation through a prompting-upon-decoding mechanism and reliable self-training via an uncertainty-driven pseudo-label calibration (UPLC) module. To facilitate research in this direction, we introduce a comprehensive ultrasound dataset spanning 5 organs and 8 segmentation tasks. Extensive experiments demonstrate that ProPL outperforms state-of-the-art methods across various metrics, establishing a new benchmark for universal ultrasound image segmentation. 
\end{abstract}

%
\begin{links}
    \link{Code}{https://github.com/WUTCM-Lab/ProPL}
\end{links}

\section{Introduction}
Medical ultrasound imaging has emerged as an indispensable diagnostic tool in clinical practice, offering real-time visualization, cost-effectiveness, and non-invasive examination capabilities. Within this context, ultrasound image segmentation plays a pivotal role in computer-aided diagnosis, surgical planning, and therapeutic assessment. However, training ultrasound image segmentation models necessitates large-scale annotated datasets, yet acquiring such datasets poses challenges. First, the annotation process is labor-intensive and time-consuming, requiring extensive expertise from medical professionals. Second, ultrasound images present inherent characteristics that make them more challenging to annotate compared to other medical imaging modalities such as CT and MRI, including speckle noise, acoustic shadows, and tissue-dependent artifacts that often obscure anatomical boundaries.
\par
Semi-supervised learning has emerged as a promising paradigm to address the scarcity of labeled data by effectively leveraging both labeled and unlabeled samples. In the context of semantic segmentation, semi-supervised approaches have demonstrated remarkable progress, primarily falling into two categories: pseudo-labeling methods and consistency-based methods. Pseudo-labeling approaches~\cite{ref_U2PL} generate pseudo targets for unlabeled data using model predictions with high confidence, subsequently utilizing these pseudo-labels for model training. Consistency-based methods~\cite{ref_unimatch}, on the other hand, enforce consistent predictions across different perturbations or views of the same input, thereby encouraging the model to learn robust representations.
\par
Despite these advances, a significant limitation persists in the field of medical image segmentation: most current approaches, whether fully supervised or semi-supervised, are designed for specific anatomical structures or diagnostic tasks. While recent works~\cite{ref_dodnet,ref_uniseg,ref_clip} have explored universal segmentation frameworks capable of handling multiple anatomical structures simultaneously, these approaches primarily rely on fully supervised learning, thereby remaining constrained by the availability of labeled data. This motivates our investigation into universal semi-supervised medical image segmentation, aiming to develop a more flexible and data-efficient approach that can generalize across different anatomical structures while requiring minimal labeled data.
\par
To this end, we introduce ProPL (prompt-guided pseudo-labeling), a framework for universal semi-supervised medical image segmentation. Our architecture comprises three key components: a shared vision encoder for generic visual feature extraction, a prompt encoder for task specification, and two independent decoders for segmentation prediction. The dual-decoder design facilitates mutual learning between the decoders through pseudo-label guidance. We propose a prompting-upon-decoding mechanism that integrates task-specific prompts with the decoders, enabling dynamic adaptation to different segmentation tasks. Furthermore, we introduce an uncertainty-driven pseudo-label calibration (UPLC) module that refines the quality of pseudo-labels by considering prediction uncertainty, thereby improving the reliability of the self-training process.
\par
To facilitate comprehensive evaluation and advance research in universal medical image segmentation, we compile and release a diverse ultrasound image dataset comprising 6,400 images spanning 5 different organs and encompassing 8 distinct segmentation tasks.
\par
Our contributions are three-fold: 
\begin{itemize}
\item We pioneer the investigation of universal semi-supervised ultrasound image segmentation, addressing a fundamental yet challenging problem in medical image analysis.
\item We propose ProPL, featuring novel components including prompting-upon-decoding and UPLC, collectively enabling universal segmentation and learning from both labeled and unlabeled data.
\item We present a comprehensive ultrasound image dataset that encompasses multiple organs and segmentation tasks. Extensive experiments demonstrate the efficacy of our approach over state-of-the-art methods.
\end{itemize}

\section{Related Works}
\subsection{Single-Task Segmentation for Ultrasound Images}
Ultrasound image segmentation has gained significant attention in recent years due to its critical role in medical diagnostics. U-Net \citep{ref_unet} establishes a seminal architecture in medical image segmentation, inspiring numerous variants tailored for ultrasound imaging. Building on this foundation, CMU-Net \citep{ref_cmunet} has been developed specifically for breast and thyroid ultrasound image segmentation tasks. Similarly, \citet{ref_asymunet} devise an asymmetric U-shaped network for breast lesion segmentation, while \citet{ref_mlfeunet} introduce a low-level feature enhancement block integrated into U-Net to improve breast lesion segmentation performance. More recently, Transformers have been explored for ultrasound segmentation. CSwin-PNet \citep{ref_cswinpnet} employs a pyramid structure with Swin Transformer as the backbone, while \citet{ref_hctnet} design a hybrid CNN-Transformer network. LM-Net \citep{ref_lmnet} presents a lightweight CNN-Transformer architecture specifically designed to enhance breast lesion segmentation in ultrasound imagery. However, training effective ultrasound image segmentation models typically requires large-scale annotated datasets, which remain challenging to obtain due to data acquisition difficulties and annotation costs.

\subsection{Single-Task Semi-Supervised Image Segmentation}
Current semi-supervised learning approaches are primarily categorized into consistency-based and pseudo-labeling methods. Among consistency-based approaches, FixMatch \citep{ref_fixmatch} generates pseudo-labels on weakly-augmented unlabeled images and trains models to predict these pseudo-labels on strongly-augmented versions of the same images. \citet{ref_augseg} propose random intensity-based augmentation combined with adaptive label-aided CutMix-based augmentation. UniMatch \citep{ref_unimatch} introduces a dual-stream augmentation strategy within the FixMatch framework to explore broader augmentation spaces. For pseudo-labeling methodologies, self-training frameworks \citep{ref_ST} utilize teacher models trained on labeled datasets to generate pseudo-labels on unlabeled images, subsequently using high-confidence pseudo-labels to supervise student models on augmented unlabeled images. U$^2$PL \citep{ref_U2PL} extends this approach by incorporating unreliable pseudo-labels into training, thereby maximizing the utilization of unlabeled data. However, existing approaches for medical images remain limited to specific organs or tasks, lacking generalizability across diverse medical imaging scenarios.

\begin{figure*}[t]
\centering\includegraphics[width=0.8\textwidth]{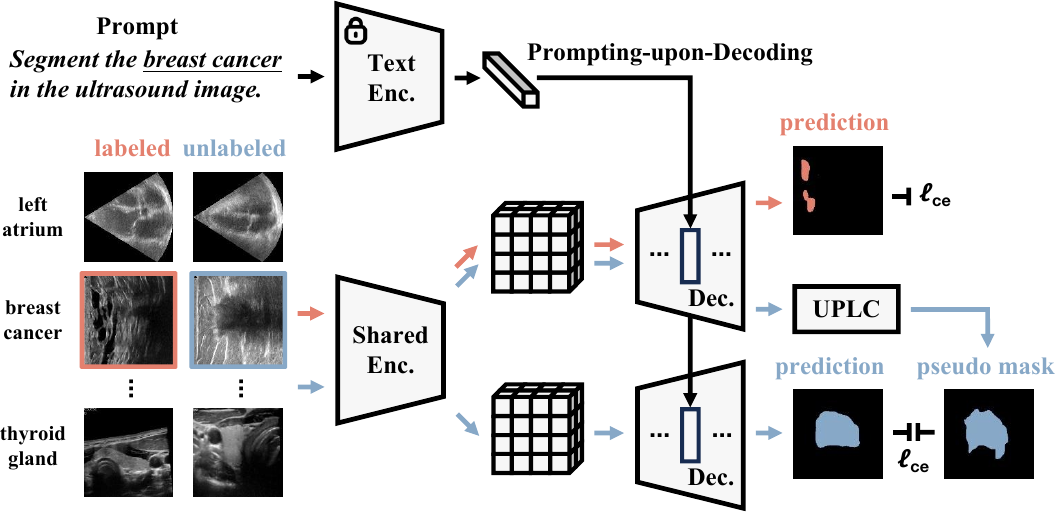}
\caption{Overview of our proposed universal semi-supervised model ProPL.}
\label{fig:chart} 
\end{figure*}

\subsection{Universal Medical Image Segmentation}
Recent advances in universal medical image segmentation focus on unifying diverse segmentation tasks, imaging modalities, and clinical objectives within single frameworks. The segment anything model (SAM) has garnered significant attention in natural image segmentation due to its outstanding performance. Several efforts have adapted SAM for medical imaging: SAMUS \citep{ref_samus} augments SAM with a two-branch CNN via cross-branch attention for ultrasound image segmentation, while DB-SAM \citep{ref_dbsam} introduces a dual-branch framework that adapts SAM for universal segmentation across 2D/3D medical images. However, SAM-based methods rely on additional cues such as point, scribble, or mask prompts, which complicates practical deployment. Alternative approaches have incorporated task priors for universal segmentation. DoDNet \citep{ref_dodnet} presents a dynamic on-demand network using a single encoder-decoder architecture with a dynamic head, encoding tasks through one-hot vectors. UniSeg \citep{ref_uniseg} substitutes the one-hot encoding in DoDNet with learnable prompts, enabling task-specific decoders. More recently, \citet{ref_clip} integrate language supervision by fusing CLIP-derived text embeddings rather than learnable prompts into the segmentation pipeline. Despite these advances in universal segmentation frameworks, these approaches remain fully supervised, thereby being constrained by labeled data availability.

\subsection{Universal Semi-Supervised Image Segmentation}
Universal semi-supervised medical image segmentation addresses challenges posed by limited labeled data and anatomical structure specificity. In computer vision for natural images, \citet{ref_usss} introduce a framework for universal semi-supervised semantic segmentation that minimizes supervised losses alongside within- and cross-domain unsupervised losses, while introducing a novel feature alignment objective based on pixel-aware entropy regularization. VerSemi \citep{ref_versemi} proposes an approach to integrate various tasks into a unified model with a broad label space, enabling utilization of large volumes of unlabeled data. To date, limited work exists on this topic, with no research focusing specifically on universal semi-supervised segmentation for ultrasound images, which represents the focus of our investigation in this paper.

\section{Methodology}
\subsection{Network Architecture}
Let $f_{\boldsymbol{\theta}}$ represent our proposed network parameterized by $\boldsymbol{\theta}$. For the $i$-th segmentation task, we define the labeled images and their corresponding ground truth masks as $\mathcal{D}^s_i = \{\bm{x}^{s}_{i,j}, \bm{y}_{i,j}\}$, where $\bm{x}^s_{i,j}$ denotes the $j$-th labeled sample. Similarly, we define unlabeled data as $\mathcal{D}^u_i = \{\bm{x}^u_{i,j}\}$, where $\bm{x}^u_{i,j}$ represents the $j$-th unlabeled example for the $i$-th task. Our ProPL framework simultaneously leverages both labeled and unlabeled data during training. As illustrated in Figure~\ref{fig:chart}, the framework consists of four key components: a shared vision encoder $\mathcal{F}_{\text{ve}}$, a prompt encoder $\mathcal{F}_{\text{pe}}$, and two decoders $\mathcal{G}_{\text{sd}}$ and $\mathcal{G}_{\text{pd}}$.
\par
The shared vision encoder $\mathcal{F}_{\text{ve}}$ in our model utilizes ConvNeXt-Tiny~\cite{ref_next}, which comprises four stages constructed using enhanced ResNet blocks. Given an input ultrasound image $\bm{x} \in \mathbb{R}^{H \times W \times 3}$, the encoder extracts hierarchical visual representations from the four stages. These multi-scale features are defined as $\bm{v}_k \in \mathbb{R}^{\frac{H}{d_k} \times \frac{W}{d_k} \times C_k}, k = 1, 2, 3, 4$, where $C_k$ denotes the number of channels at stage $k$, and $H$ and $W$ represent the height and width of the input image, respectively.
\par
Each of our dual decoders consists of four stages, with skip connections between the encoder and decoders implemented following UNETR~\cite{ref_unetr}. In addition, sub-pixel convolutions are employed for upsampling. Similar to the encoder features $\bm{v}_k$, we denote the corresponding features at each decoding stage as $\bm{z}_k \in \mathbb{R}^{\frac{H}{d_k} \times \frac{W}{d_k} \times C_k}$.
\par
For task-specific prompting, we leverage BERT~\cite{ref_bert} to extract textual features $\bm{t} \in \mathbb{R}^{L \times D}$ from a given task prompt, where $D$ represents the feature dimension and $L$ denotes the length of the prompt. This encoder transforms task-specific prompts into continuous representations that the model can interpret, enabling the model to comprehend ongoing segmentation tasks via our proposed prompting-upon-decoding module.

\subsection{Prompting-upon-Decoding}
\label{subsec:pud}

To incorporate task-specific information, we integrate $\bm{t}$ with the decoders. First, we perform self-attention on $\bm{z}_k$:
\begin{equation}\label{eq:mhsa}
\bm{z}_k^\prime = \bm{z}_k + \mathrm{LN}(\mathrm{MHSA}(\bm{z}_k)) \,,
\end{equation}
where $\mathrm{MHSA}(\cdot)$ represents multi-head self-attention, and $\mathrm{LN}(\cdot)$ denotes layer normalization.
\par
Subsequently, $\bm{t}$ is processed through a 1D convolution, followed by a linear projection to align the dimension of textual features with that of the visual representation $\bm{z}_k$:
\begin{equation}\label{eq:conv}
\boldsymbol{\tau} = \mathrm{Conv}(\bm{t})\mathbf{W} \,,
\end{equation}
where $\mathbf{W}$ is a learnable mapping matrix. $\boldsymbol{\tau}\in\mathbb{R}^{M \times C_i}$ is the resulting text embedding.
\par
Finally, we employ multi-head cross-attention (MHCA) to inject the prompt information into our model:
\begin{equation}
\bm{h}_k = \bm{z}_k^\prime + \alpha \mathrm{LN}(\mathrm{MHCA}(Q = \bm{z}_k^\prime, K = \boldsymbol{\tau}, V = \boldsymbol{\tau})) \,,
\end{equation}
where $\alpha$ is a learnable parameter that modulates the influence of the prompt features.
\begin{table*}[h!]
  \centering
  \large
  \def\arraystretch{1.2}
  
    \resizebox{1\textwidth}{!}{
    \begin{tabular}{p{0.4cm}|p{2.6cm}|cc|cc|cc|cc|cc|cc|cc|cc|c|c}
\hline
    \multicolumn{2}{@{\hspace{-0.95cm}}c|}{\multirow{2}{*}{\raisebox{-0.5pt}{\textbf{Method}}}} &\multicolumn{2}{c|}{\raisebox{-0.5pt}{\textbf{BC}}} &\multicolumn{2}{c|}{\raisebox{-0.5pt}{\textbf{FH}}} &\multicolumn{2}{c|}{\raisebox{-0.5pt}{\textbf{LA}}} & \multicolumn{2}{c|}{\raisebox{-0.5pt}{\textbf{LV}}} &\multicolumn{2}{c|}{\raisebox{-0.5pt}{\textbf{MD}}} & \multicolumn{2}{c|}{\raisebox{-0.5pt}{\textbf{OT}}}  & \multicolumn{2}{c|}{\raisebox{-0.5pt}{\textbf{TG}}} & \multicolumn{2}{c|}{\raisebox{-0.5pt}{\textbf{TN}}} & \multicolumn{1}{c|}{\multirow{2}{*}{\raisebox{-0.5pt}{\textbf{mDice}}}} & \multicolumn{1}{c}{\multirow{2}{*}{\raisebox{-0.5pt}{\textbf{mIoU}}}} \\
     \multicolumn{2}{@{\hspace{-0.95cm}}c|}{ } &\textbf{Dice} &\textbf{IoU} &\textbf{Dice} &\textbf{IoU} &\textbf{Dice} &\textbf{IoU} &\textbf{Dice} &\textbf{IoU} &\textbf{Dice} &\textbf{IoU} &\textbf{Dice} &\textbf{IoU} &\textbf{Dice} &\textbf{IoU} &\textbf{Dice} &\textbf{IoU} & &\\
\hline
     \multicolumn{1}{c|}{\multirow{10}{*}{\ding{171}}} &U-Net & 64.15 &52.25 & 94.76 &90.44 & 80.07 &69.78 & 87.26 &78.00 & 76.80 &63.08  & 67.96 &56.48 & 78.07 &67.89 & 52.27 &40.18 & 75.17 &64.76 \\
     &PSPNet & 64.56 &52.45 	&93.13 	&87.96 	&76.27 	&65.04 	&86.23 	&76.57 	&73.80 	&59.43 	&60.64 	&48.44 	&75.89 	&65.22 &51.84 	&39.97 	&72.80 	&61.89  \\
     &DeepLabv3+ &65.90 &54.21 	&94.58 	&90.07 	&74.43 	&62.47 	&84.98 &74.78 	&71.25 	&56.37 	&65.48 	&53.56 	&76.20 	&65.10 	&56.00 	&44.00 	&73.60 	&62.57 
 \\
     &UNet++   &50.49 	&38.75 	&87.18 	&79.12 	&71.45 	&58.17 	&82.53 	&71.43 	&68.04 	&52.71 	&57.17 	&43.46 	&72.06 	&60.52 	&43.48 	&30.67 	&66.55 	&54.35 
 \\
     &SegFormer &70.85 	&59.30 	&95.02 	&90.89 	&74.59 	&62.39 	&86.41 	&76.66 	&72.11 	&57.07 	&69.90 	&58.14 	&76.73 	&66.34 	&54.76 	&42.01 	&75.05 	&64.10 
 \\
     &Swin-Unet & 65.63 	&53.39 	&95.35 	&91.33 	&77.06 	&65.03 	&85.20 	&74.83 	&67.69 	&51.82 	&67.97 	&56.22 	&75.28 	&63.80 	&51.19 	&38.53 	&73.17 	&61.87 
 \\
    &Swin UNETR &53.81 	&41.05 	&86.54 	&78.34 	&69.37 	&56.63 	&84.19 	&73.50 	&68.74 	&53.34 	&60.78 	&48.18 &74.12 &62.59 &40.43 &27.53 &67.25 &55.15 
 \\
     &HiFormer &55.93 &43.17 &92.97 &87.52 &74.26 &61.78 &86.33 &76.56 &72.58 &57.75 &64.70 &	51.66 &77.95 &67.10 &52.19 &38.43 &72.11 &60.50 
  \\
    &H2Former  &41.20 &27.89 &85.03 &75.60 &55.67 &40.43 &82.34 &70.84 &55.73 &39.29 &60.77 	&47.70 &68.24 &56.28 &26.91 &16.58 &59.49 &46.83 
 \\
    &Mamba-UNet  & 65.38 &53.73 &\textbf{96.54} &93.47 &78.47 &68.47 &86.92 &77.51 &77.66 &64.16 &62.74 	&52.76 &78.85 &69.42 &61.03 &49.18 &75.95 &66.09 
 \\
\hline 
    \multicolumn{1}{c|}{\multirow{7}{*}{\ding{170}}} &FixMatch   & 33.92 &26.23 &87.92 &79.51 	&26.10 &16.74 &78.35 &65.60 &63.64 &47.65 &33.85 &22.17 &73.29 &63.06 &29.86 &22.13 &53.37 	&42.89 
  \\
    &U$^2$PL   & 64.48 &52.46 &93.83 &89.05 &71.58 &57.75 &83.52 &72.63 &70.99 &56.21 &67.87 &55.18 &78.68 	&68.04 &49.27 &37.97 &72.53 &61.16 
  \\
    &ST++  & 39.85 &29.38 &83.12 &72.52 &60.85 &47.39 &71.54 &58.16 &56.61 &42.37 &40.56 &31.27 	&66.82 &54.35 &31.33 &21.92 &56.34 &44.67 
 \\
    &UniMatch   & 43.36 &33.41 &87.64 &78.99 &34.54 &23.05 &79.25 &66.68 &58.30 &42.35 &67.88 &55.17 &74.68 &64.49 &51.71 &39.64 &62.17 &50.47 
\\
    &AugSeg & 15.66 &10.15 &82.27 &71.92 &18.88 &11.73 &50.94 & 35.20 &36.17 &23.04 &27.06 &18.15 &41.72 &28.16 &21.11 &13.73 &36.73 &26.51

\\
    &DDFP & 71.86 &58.72 &95.85 &\textbf{94.82} &85.05 &75.37 &88.74 &80.68 &77.48 &62.03 &73.04 &\textbf{68.43} &82.68 &\textbf{79.17} &66.56 &\textbf{61.99} &80.16 &\textbf{72.65}

 \\
    &CSC-PA &49.04 &36.97 &87.16 &79.84 &76.06 &63.72 &83.66 &72.70 &74.16 &59.92 &60.96 &48.90 	&78.60 &68.26 &46.79 &34.70 &69.55 &58.13 \\

\hline
    \multicolumn{1}{c|}{\multirow{2}{*}{\ding{168}}} &DoDNet   &73.06 &57.55 &94.94 &90.37 &29.79 	&17.50 &61.97 &44.90 &19.44 &10.77 &79.42 &65.87 &\textbf{85.59} &74.82 &65.93 &49.17 &63.77 &51.37 
 \\
    &CLIP-UM  & 71.20 &55.28 &95.21 &90.87 &43.20 &27.55 &36.54 &22.35 &23.96 &13.61 &\textbf{80.20} &66.95 	&85.13 &74.11 &\textbf{69.93} &53.76 &63.17 &50.56 
\\
\hline
    \multicolumn{1}{c|}{\multirow{2}{*}{\ding{169}}} &Univ-full & 43.68 &32.59 &86.07 &77.49 	&21.89 &12.76 &43.88 &30.23 &30.55 &18.62 &52.37 &39.26 &57.12 &45.05 &43.58 &31.80 &47.39 	&35.98 
 \\
    &ProPL  & \textbf{73.71} &\textbf{63.76} &95.65 &92.17 &\textbf{86.63} &\textbf{77.58} &\textbf{91.11} &\textbf{83.94} &\textbf{81.31} &\textbf{69.05} &73.49 &63.83 &81.21 &71.14 &65.96 &54.22 &\textbf{81.13} &71.96 
 \\
\hline
    \end{tabular}%
    }
  \caption{Performance comparison of single-task (supervised \ding{171}, semi-supervised \ding{170}) and universal (supervised \ding{168}, semi-supervised \ding{169}) models across eight ultrasound image segmentation tasks under the 1/16 data partition. Metrics reported as Dice (\%) per task, mean Dice (mDice) and mean IoU (mIoU). Best results in \textbf{bold}. Tasks: breast cancer (BC), fetal head (FH), left atrium (LA), left ventricle (LV), myocardium (MD), ovarian tumor (OT), thyroid gland (TG), and thyroid nodule (TN).}
  \label{tab:comparison116}%
\end{table*}

    
    

\begin{table*}[h!]
  \centering
  \large
  \def\arraystretch{1.2}
  
    \resizebox{1\textwidth}{!}{
    \begin{tabular}{p{0.4cm}|p{2.6cm}|cc|cc|cc|cc|cc|cc|cc|cc|c|c}
\hline
    \multicolumn{2}{@{\hspace{-0.95cm}}c|}{\multirow{2}{*}{\raisebox{-0.5pt}{\textbf{Method}}}} &\multicolumn{2}{c|}{\raisebox{-0.5pt}{\textbf{BC}}} &\multicolumn{2}{c|}{\raisebox{-0.5pt}{\textbf{FH}}} &\multicolumn{2}{c|}{\raisebox{-0.5pt}{\textbf{LA}}} & \multicolumn{2}{c|}{\raisebox{-0.5pt}{\textbf{LV}}} &\multicolumn{2}{c|}{\raisebox{-0.5pt}{\textbf{MD}}} & \multicolumn{2}{c|}{\raisebox{-0.5pt}{\textbf{OT}}}  & \multicolumn{2}{c|}{\raisebox{-0.5pt}{\textbf{TG}}} & \multicolumn{2}{c|}{\raisebox{-0.5pt}{\textbf{TN}}} & \multicolumn{1}{c|}{\multirow{2}{*}{\raisebox{-0.5pt}{\textbf{mDice}}}} & \multicolumn{1}{c}{\multirow{2}{*}{\raisebox{-0.5pt}{\textbf{mIoU}}}} \\
     \multicolumn{2}{@{\hspace{-0.95cm}}c|}{ } &\textbf{Dice} &\textbf{IoU} &\textbf{Dice} &\textbf{IoU} &\textbf{Dice} &\textbf{IoU} &\textbf{Dice} &\textbf{IoU} &\textbf{Dice} &\textbf{IoU} &\textbf{Dice} &\textbf{IoU} &\textbf{Dice} &\textbf{IoU} &\textbf{Dice} &\textbf{IoU} & &\\
\hline
     \multicolumn{1}{c|}{\multirow{10}{*}{\ding{171}}} &U-Net & 71.45 &60.47 &96.00 &92.54 &84.24 &74.56 &89.34 &81.09 &79.45 &66.68 &76.10 &65.34 &81.36 &71.84 &61.74 &49.64 &79.96 &70.27 
 \\
     &PSPNet & 70.14 &58.54 &94.89 &90.80 &82.22 &71.92 &87.96 &79.12 &78.27 &65.01 &70.46 &58.42 &79.82 &70.36 &57.37 &45.64 &77.64 &67.48 
  \\
     &DeepLabv3+ &72.04 &61.13 &95.46 &91.62 &81.93 &71.42 &87.23 &78.02 &77.08 &63.44 &73.87 &62.62 &78.79 &69.05 &58.77 &46.54 &78.15 &67.98 
 
 \\
     &UNet++   &53.68 &41.60 &89.16 &82.08 &74.20 &61.98 &86.35 &76.61 &72.09 &57.29 &62.25 &49.27 &76.89 &65.87 &45.61 &32.42 &70.03 &58.39 

 \\
     &SegFormer &76.08 &65.34 &95.89 &92.35 &80.31 &69.21 &87.58 &78.44 &76.31 &62.37 &76.37 &65.44 &78.56 &68.75 &63.02 &50.85 &79.27 &69.09 
 
 \\
     &Swin-Unet & 74.29 &63.17 &96.11 &92.60 &82.04 &71.03 &87.85 &78.71 &74.80 &60.37 &75.14 &64.05 &78.68 &68.36 &63.25 &50.84 &79.02 &68.64 
 
 \\
    &Swin UNETR &59.21 &46.99 &89.72 &83.10 &77.48 &66.01 &86.43 &76.80 &73.58 &59.13 &64.92 &51.97 &78.70 &68.34 &46.32 &33.50 &72.05 &60.73 
 
 \\
     &HiFormer &62.80 &50.92 &95.32 &91.35 &80.41 &69.40 &88.15 &79.30 &76.89 &63.03 &71.14 &59.32 &79.94 &69.87 &58.84 &46.03 &76.69 &66.15 
 
  \\
    &H2Former  &59.23 &46.96 &93.23 &88.07 &75.87 &63.78 &87.13 &77.77 &62.49 &46.11 &64.33 &51.21 	&72.89 &61.29 &48.14 &35.29 &70.41 &58.81 
 
 \\
    &Mamba-UNet  & 73.78 &63.86 &96.71 &93.78 &83.30 &73.16 &88.13 &79.41 &79.99 &67.38 &78.04 &68.22 &83.20 &75.25 &66.06 &54.35 &81.15 &71.93 
 
 \\
    
\hline 
    \multicolumn{1}{c|}{\multirow{7}{*}{\ding{170}}} &FixMatch   & 55.54 & 43.89 & 88.47 & 80.42 &38.56 &26.68 &80.52 &68.12 &63.10 &47.15 &36.46 &25.13 &75.22 &65.55 &42.41 &32.22 &60.04 &48.65 

  \\
    &U$^2$PL   & 62.19 &50.78 &95.10 &91.10 &77.27 &64.73 &86.33 &76.60 &76.71 &63.06 &70.91 &59.08 &80.93 &71.22 &54.43 &42.48 &75.48 &64.88 
 
  \\
    &ST++  & 51.08 &39.76 &88.24 &80.83 &64.02 &51.48 &79.21 &66.82 &65.72 &50.75 &58.03 &46.85 &72.83 &61.03 &43.44 &31.73 &65.32 &53.66 
 
 \\
    &UniMatch   & 58.65 &46.63 &88.04 &79.48 &39.34 &27.29 &78.04 &64.97 &60.68 &44.52 &64.83 &51.24 &75.06 &65.26 &56.04 &43.36 &65.09 &52.84
\\
    &AugSeg & 14.74 &9.64 &85.45 &75.95 &43.24 &29.57 &58.94 &42.51 &29.55	&18.13 &43.63 &31.96 &46.03 &32.30 &22.36 &14.90 &42.99	&31.87

\\
    &DDFP & 72.54 &62.50  &\textbf{97.07} &\textbf{95.13} &85.72 &76.57 &90.23 &83.32 &80.51 &66.86 &76.54 &71.78	&84.07 &80.32 &68.81 &\textbf{63.52}                               &81.94 &\textbf{75.00}

 \\
    &CSC-PA &49.04 &36.97 &87.16 &79.84 &76.06 &63.72 &83.66 &72.70 &74.16 &59.92 &60.96 &48.90 	&78.60 &68.26 &46.79 &34.70 &69.55 &58.13 
    \\
\hline
    \multicolumn{1}{c|}{\multirow{2}{*}{\ding{168}}} &DoDNet   &78.15 &64.14 &96.73 &93.67 &30.50 	&17.99 &43.80 &28.04 &25.34 &14.51 &82.81 &70.67 &\textbf{90.18} &\textbf{82.11} &74.68 &	59.59 &65.27 &53.84 
 
 \\
    &CLIP-UM  & \textbf{79.17} &\textbf{65.52} &96.49 &93.22 &17.84 &9.79 &64.22 &47.30 &43.86 &28.09 &\textbf{84.51} &\textbf{73.18} 	&89.58 &81.13 &\textbf{76.18} &61.53 &68.98 &57.47 
 
\\
\hline
    \multicolumn{1}{c|}{\multirow{2}{*}{\ding{169}}} &Univ-full & 48.87 &37.31 &85.13 &75.87 	&19.76 &11.22 &58.51 &43.11 &19.97 &11.27 &56.66 &45.41 &56.99 &46.04 &44.62 &33.22 &48.81 	&37.93 
 
 \\
    &ProPL  & 75.03 &65.29 &96.25 &93.20 &\textbf{87.16} &\textbf{78.85} &\textbf{92.29} &\textbf{85.90} &\textbf{84.13} &\textbf{73.08} &77.05 &66.91 	&83.61 &75.59 &71.30 &59.83 &\textbf{83.35} &74.83 
 
 \\
\hline
    \end{tabular}%
    }
  \caption{Performance comparison of single-task (supervised \ding{171}, semi-supervised \ding{170}) and universal (supervised \ding{168}, semi-supervised \ding{169}) models across eight ultrasound image segmentation tasks under the 1/8 data partition.}
  \label{tab:comparison18}%
\end{table*}

\begin{table*}[h!]
  \centering
  \large
  \def\arraystretch{1.2}
  
    \resizebox{1\textwidth}{!}{
    \begin{tabular}{p{0.4cm}|p{2.6cm}|cc|cc|cc|cc|cc|cc|cc|cc|c|c}
\hline
    \multicolumn{2}{@{\hspace{-0.95cm}}c|}{\multirow{2}{*}{\raisebox{-0.5pt}{\textbf{Method}}}} &\multicolumn{2}{c|}{\raisebox{-0.5pt}{\textbf{BC}}} &\multicolumn{2}{c|}{\raisebox{-0.5pt}{\textbf{FH}}} &\multicolumn{2}{c|}{\raisebox{-0.5pt}{\textbf{LA}}} & \multicolumn{2}{c|}{\raisebox{-0.5pt}{\textbf{LV}}} &\multicolumn{2}{c|}{\raisebox{-0.5pt}{\textbf{MD}}} & \multicolumn{2}{c|}{\raisebox{-0.5pt}{\textbf{OT}}}  & \multicolumn{2}{c|}{\raisebox{-0.5pt}{\textbf{TG}}} & \multicolumn{2}{c|}{\raisebox{-0.5pt}{\textbf{TN}}} & \multicolumn{1}{c|}{\multirow{2}{*}{\raisebox{-0.5pt}{\textbf{mDice}}}} & \multicolumn{1}{c}{\multirow{2}{*}{\raisebox{-0.5pt}{\textbf{mIoU}}}} \\
     \multicolumn{2}{@{\hspace{-0.95cm}}c|}{ } &\textbf{Dice} &\textbf{IoU} &\textbf{Dice} &\textbf{IoU} &\textbf{Dice} &\textbf{IoU} &\textbf{Dice} &\textbf{IoU} &\textbf{Dice} &\textbf{IoU} &\textbf{Dice} &\textbf{IoU} &\textbf{Dice} &\textbf{IoU} &\textbf{Dice} &\textbf{IoU} & &\\
\hline
     \multicolumn{1}{c|}{\multirow{10}{*}{\ding{171}}} &U-Net & 78.55 &68.75 &96.66 &93.66 &86.32 &77.48 &90.30 &82.78 &82.43 &70.73 &79.77 &70.29 &86.51 &78.42 &69.64 &58.62 &83.77 &75.09 
 
 \\
     &PSPNet & 76.21 &65.58 &96.39 &93.18 &85.18 &76.40 &90.11 &82.34 &81.76 &69.71 &76.65 &66.10 &84.78 &78.89 &66.23 &54.26 &82.16 &73.31 
 
  \\
     &DeepLabv3+ &76.78 &66.82 &96.46 &93.32 &83.84 &74.54 &89.64 &81.58 &80.59 &68.09 &79.20 &69.11 &85.48 &76.91 &67.04 &55.48 &82.38 &73.23

 \\
     &UNet++   &61.56 &49.22 &92.11 &86.69 &82.36 &72.39 &88.59 &79.97 &77.69 &64.32 &65.42 &53.00 &83.25 &73.05 &50.04 &37.07 &75.13 &64.46

 \\
     &SegFormer &78.35 &68.36 &96.37 &93.19 &83.80 &74.33 &89.54 &81.46 &79.86 &66.98 &78.98 &68.61 &84.44 &75.35 &71.15 &58.95 &82.81 &73.40

 \\
     &Swin-Unet & 77.66 &67.19 &96.41 &93.14 &84.99 &75.22 &87.19 &77.74 &78.74 &65.55 &78.69 &68.00 &82.43 &72.74 &71.07 &58.90 &82.15 &72.31

 \\
    &Swin UNETR &65.53 &53.83 &93.15 &88.21 &83.21 &73.10 &88.29 &79.55 &78.39 &65.23 &67.68 &55.35 &84.26 &75.35 &50.53 &37.84 &76.38 &66.06

 \\
     &HiFormer &73.82 &63.40 &96.09 &92.66 &83.90 &74.42 &89.33 &81.20 &80.15 &67.37 &76.33 &65.19 &83.53 &74.57 &65.45 &53.16 &81.08 &71.50 
 
  \\
    &H2Former  &70.78 &59.47 &95.15 &91.19 &83.96 &74.56 &89.00 &80.84 &77.78 &64.18 &74.82 &63.03 &81.41 &72.14 &58.64 &46.14 &78.94 &68.94

 \\
    &Mamba-UNet  & 78.59 &\textbf{69.59} &97.06 &94.38 &85.22 &76.38 &89.87 &82.00 &82.23 &70.41 &81.87 	&72.47 &86.99 &79.26 &71.57 &60.60 &84.18 &75.64 
 
 \\
    
\hline 
    \multicolumn{1}{c|}{\multirow{7}{*}{\ding{170}}} &FixMatch   & 56.00 &44.37 &84.99 &75.70 	&43.88 &30.57 &81.99 &70.36 &62.79 &47.19 &64.92 &51.57 &76.06 &66.29 &55.27 &43.46 &65.74 	&53.69

  \\
    &U$^2$PL   & 69.16 &57.80 & 95.73 & 92.10 & 80.61 & 69.06 & 88.14 & 79.36 & 78.36 & 65.19 & 72.14 &60.40 &82.83 &73.68 &59.24 &46.47 &78.28 &68.01

  \\
    &ST++  & 58.47 &46.33 &92.58 &87.30 &66.34 &52.22 &81.58 &70.16 &70.99 &55.84 &64.66 &53.14 &76.04 &64.26 &47.63 &35.65 &69.79 &58.11

 \\
    &UniMatch   & 60.95 &48.82 &88.63 &80.60 &42.65 &29.91&	79.96 &67.60 &63.92 &48.17 &67.88 &55.17 &76.32 &66.54 &59.50 &46.97 &67.48 &55.47 
\\
    &AugSeg & 35.07 &25.60 &88.28 &80.23 &48.25 &32.79 &63.57 &47.54 &20.01 &11.72 &47.25 &33.54 &43.54 &30.21 &36.56 &26.51 &47.82 &36.02

 \\
    &DDFP &71.86 &58.72	&96.85 &\textbf{94.82} &85.95 &75.37	&88.74 &80.68 &77.48 &62.03 &73.04 &68.43 &82.68 &79.17 &66.56 &61.99 &80.40 &72.65

 \\
    &CSC-PA &49.91 &37.38 &89.05 &82.16 &75.62 &63.21 &84.69 &74.12 &74.41 &60.08 &60.95 &48.44 &78.37 &67.90 &45.51 &33.21 &69.81 &58.31 
 
    \\
\hline
    \multicolumn{1}{c|}{\multirow{2}{*}{\ding{168}}} &DoDNet   &\textbf{79.72} &66.28 &\textbf{97.08} &94.33 &34.87 	&21.11 &58.29 &41.13 &75.69 &60.88 &\textbf{85.86} &\textbf{75.22} &\textbf{92.47} &\textbf{86.00} &77.67 &63.47 &75.21 &63.55

 \\
    &CLIP-UM  & 79.21 &65.57 &96.58 &93.39 &84.71 &73.48 &75.43 &60.55 &71.25 &55.34 &85.49 &74.66 	&88.95 &80.11 &\textbf{77.69} &63.52 &82.41 &70.83

\\
\hline
    \multicolumn{1}{c|}{\multirow{2}{*}{\ding{169}}} &Univ-full & 55.71 &43.52 &85.81 &76.71 	&38.66 &25.05 &53.10 &37.33 &39.89 &26.65 &62.60 &51.05 &63.77 &52.12 &46.84 &35.75 &55.80 	&43.52

 \\
    &ProPL  & 77.18 &67.91 &96.56 &93.64 &\textbf{88.72} &\textbf{80.83} &\textbf{93.00} &\textbf{87.07} &\textbf{85.24} &\textbf{74.70} &78.31 &68.86 	&89.52 &82.14 &75.44 &\textbf{64.50} &\textbf{85.50} &\textbf{77.46}

 \\
\hline
    \end{tabular}%
    }
  \caption{Performance comparison of single-task (supervised \ding{171}, semi-supervised \ding{170}) and universal (supervised \ding{168}, semi-supervised \ding{169}) models across eight ultrasound image segmentation tasks under the 1/4 data partition.}
  \label{tab:comparison14}%

\end{table*}

\subsection{Training and Pseudo-Label Calibration}

We present the training of our pseudo-labeling-based framework for universal semi-supervised ultrasound image segmentation, followed by the proposed UPLC mechanism that enhances pseudo-label quality.

\noindent
\textbf{Pseudo-Labeling}
For each labeled image $\bm{x}_{i,j}^s \in \mathcal{D}_i^s$, we process both the image through the vision encoder $\mathcal{F}_{\text{ve}}$ and its corresponding task prompt through the prompt encoder $\mathcal{F}_{\text{pe}}$. The decoder $\mathcal{G}_{\text{sd}}$, incorporating the prompting-upon-decoding module, then generates a segmentation mask. We optimize this process using a combined loss function that uses binary cross-entropy and Dice losses to minimize the discrepancy between the predicted and ground truth masks.
\par
For an unlabeled image $\bm{x}_{i,j}^u \in \mathcal{D}_i^u$, we first extract feature representations using $\mathcal{F}_{\text{ve}}$ and generate a pseudo segmentation mask through $\mathcal{G}_{\text{sd}}$. These features are subsequently processed by $\mathcal{G}_{\text{pd}}$ to predict another segmentation mask. We apply the same combined loss function to enforce consistency between these two masks.

\begin{figure*}[h!]
\centering\includegraphics[width=0.75\textwidth]{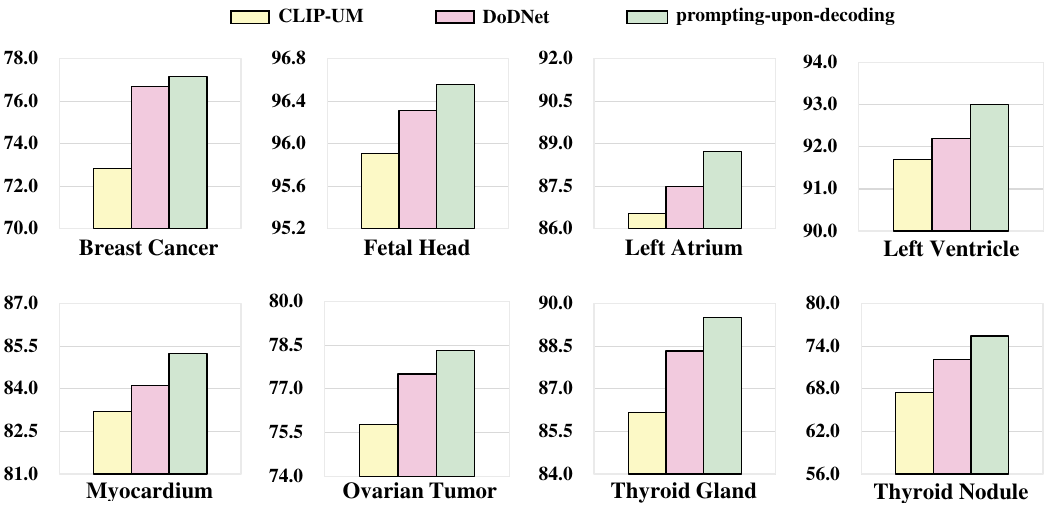}
\caption{Comparison of our prompting-upon-decoding method against established prompting approaches implemented in CLIP-UM~\cite{ref_clip} and DoDNet~\cite{ref_dodnet}.}
\label{fig:PUD}
\end{figure*}

\begin{figure*}[h!]
\begin{minipage}{0.5\textwidth}
 \centerline{\includegraphics[width=0.9\textwidth]{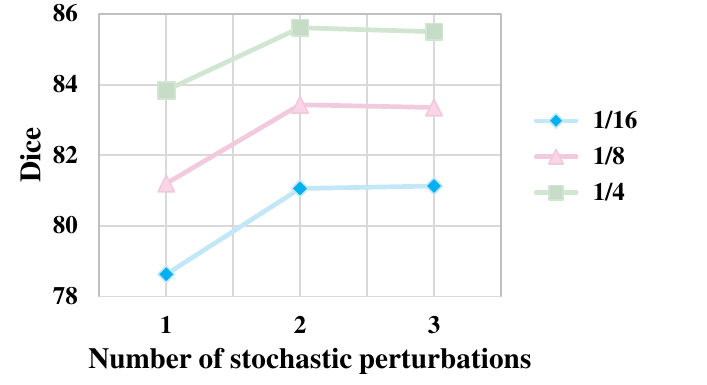}}
 \caption{Impact of varying the number of stochastic perturbations in UPLC.}
 \label{fig:passtime}
\end{minipage}
\hspace{0.2cm}
\begin{minipage}{0.5\textwidth}
 \centerline{\includegraphics[width=0.9\textwidth]{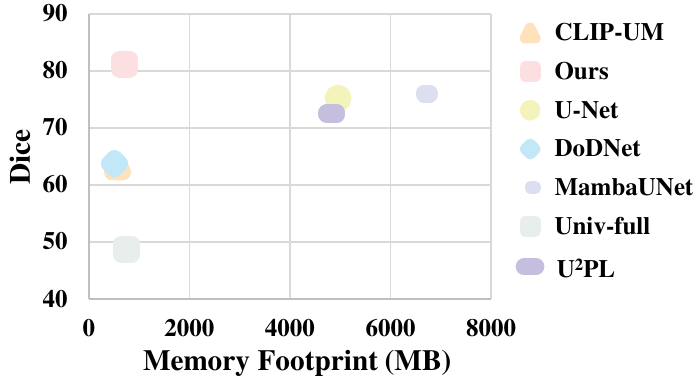}}
 \caption{Performance-efficiency trade-off analysis across various models.}
 \label{fig:memory}
\end{minipage}
\end{figure*}

\noindent
\textbf{UPLC}
While pseudo-labels are valuable for semi-supervised learning, they often contain inherent noise that can impact model performance. To address this limitation, we introduce UPLC for pseudo-label refinement.

\begin{figure*}[h!]
\centering\includegraphics[width=0.8\textwidth]{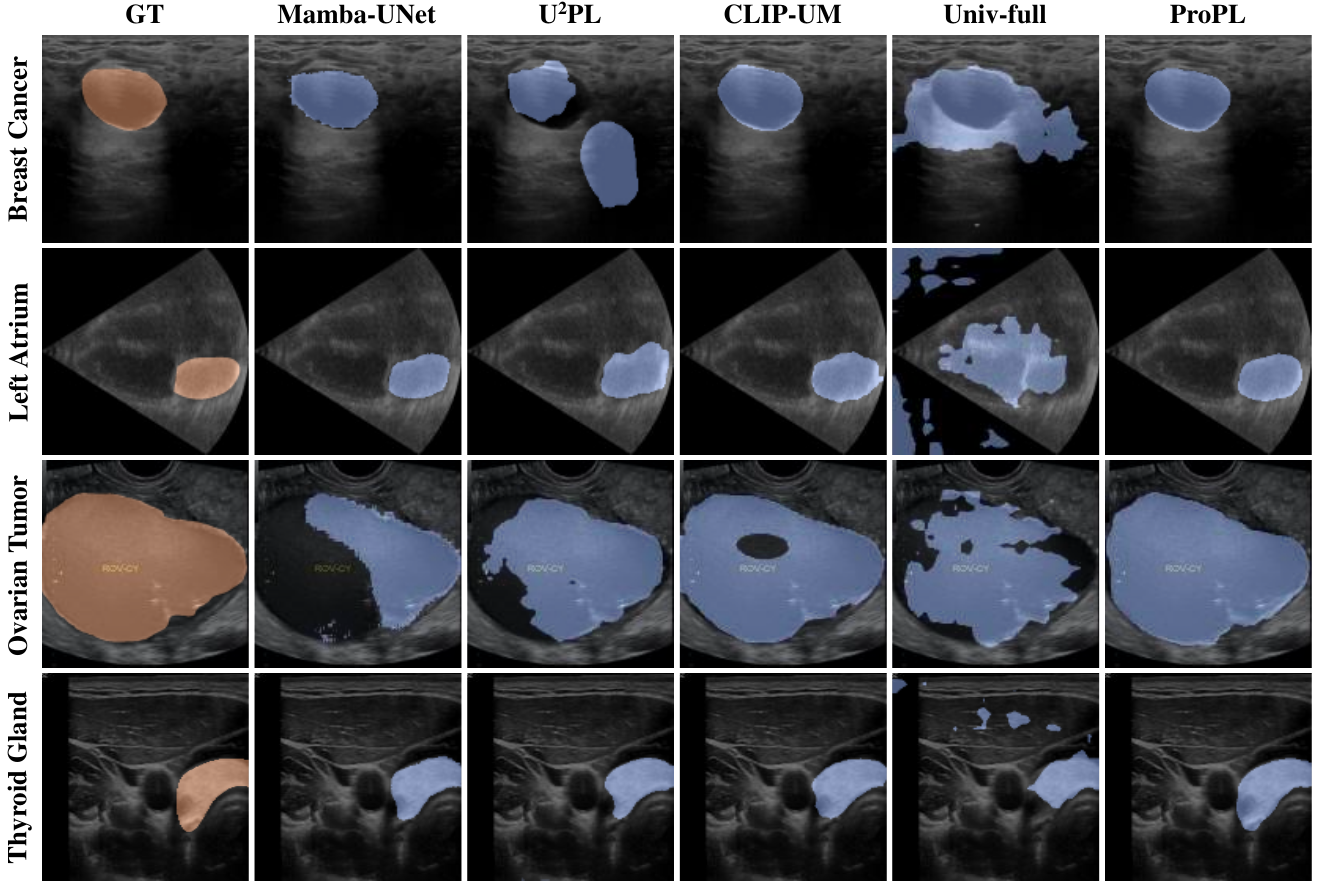}
\caption{Qualitative segmentation results comparing the proposed method against competitive approaches across four segmentation tasks: breast cancer, left atrium, ovarian tumor, and thyroid gland.}
\label{fig:vision} 
\end{figure*}

\par
Given an unlabeled image $\bm{x}^u$ (omitting subscripts $i$ and $j$ for notational simplicity), we apply $N$ stochastic perturbations to its encoded visual representation $\bm{v}_4$ before processing through $\mathcal{G}_{\text{sd}}$. This generates an ensemble of pseudo segmentation masks $\{\hat{\bm{y}}_i\}$. We then compute the empirical mean and variance of these masks:
\begin{equation}
\label{eq:uplc_1}
\boldsymbol{\mu} = \frac{1}{N} \sum_{i=1}^{N} \hat{\bm{y}}_i \,, \quad
\boldsymbol{\gamma} = \frac{1}{N} \sum_{i=1}^{N} (\hat{\bm{y}}_i - \boldsymbol{\mu})^2 \,.
\end{equation}
\par
The variance $\boldsymbol{\gamma}$ provides an uncertainty measure for the pseudo-labels. We leverage this uncertainty to modulate supervision signals for $\mathcal{G}_{\text{pd}}$ by defining rectified pseudo-labels as:
\begin{equation}
\hat{\bm{y}} = \exp(-\boldsymbol{\gamma}) \odot \boldsymbol{\mu} \,,
\end{equation}
where $\odot$ denotes element-wise multiplication.

\section{Experiments}

\subsection{Datasets and Evaluation Metrics}
We evaluate our network on eight ultrasound image segmentation tasks spanning 6,400 samples, with 800 samples per task. Our dataset comprises breast cancer images from Dataset B \cite{ref_datasetb} and BUSI \cite{ref_busi}, fetal head images from HC18 \cite{ref_hc18}, left atrium and left ventricle data from CAMUS \cite{ref_camus}, myocardium data from CAMUS and HMC-QU \cite{ref_hmcqu}, ovarian tumor images from MMOTU \cite{ref_mmotu}, thyroid gland images from TN3K \cite{ref_tn3k}, and thyroid nodule data from TN3K and DDTI \cite{ref_ddti}. We maintain a 3:1 train-test split ratio across all tasks. 
\par
Following standard semi-supervised learning protocols, we conduct experiments using 1/16, 1/8, and 1/4 of the training data as labeled samples, with the remainder serving as unlabeled data. We evaluate performance using the Dice similarity coefficient and mean Intersection over Union (mIoU).


\subsection{Implementation Details}
We implement our framework in PyTorch, conducting all experiments on a single NVIDIA GeForce RTX 4090 GPU. All the images are resized to 224$\times$224 pixels and augmented through random rotation and scaling. We train with a batch size of 16 using SGD ($\text{momentum}=0.9$, $\text{weight decay}=1e-5$). The learning rate follows a polynomial schedule: $lr = init_{lr} \times (1- \frac{iter}{max_{iter}})^{power}$ where $init_{lr} = 0.001$ and $power = 0.9$. Models are trained for 200 epochs. We train the model for 200 epochs. Within UPLC, we introduce perturbations via dropout (rate 0.3) and draw two stochastic perturbation instances.

For text prompts, we use minimal instructions such as ``Segment the breast cancer in the ultrasound image.'' The term ``breast cancer'' can be replaced with other targets, including the left atrium, ovarian tumor, or thyroid gland.
\begin{table*}[h!]
  \centering
  \large
  \def\arraystretch{1.3}
  
    \resizebox{1\textwidth}{!}{
    \begin{tabular}{>{\centering}m{1.2cm} >{\centering}m{1.2cm}|cc|cc|cc|cc|cc|cc|cc|cc|c|c}
\hline
    \multirow{2}{*}{\raisebox{-0.5pt}{\textbf{PuD}}} &\multirow{2}{*}{\raisebox{-0.5pt}{\textbf{UPLC}}} &\multicolumn{2}{c|}{\raisebox{-0.5pt}{\textbf{BC}}} &\multicolumn{2}{c|}{\raisebox{-0.5pt}{\textbf{FH}}} &\multicolumn{2}{c|}{\raisebox{-0.5pt}{\textbf{LA}}} & \multicolumn{2}{c|}{\raisebox{-0.5pt}{\textbf{LV}}} & \multicolumn{2}{c|}{\raisebox{-0.5pt}{\textbf{MD}}} & \multicolumn{2}{c|}{\raisebox{-0.5pt}{\textbf{OT}}} & \multicolumn{2}{c|}{\raisebox{-0.5pt}{\textbf{TG}}} & \multicolumn{2}{c|}{\raisebox{-0.5pt}{\textbf{TN}}} & \multirow{2}{*}{\raisebox{-0.5pt}{\textbf{mDice}}} & \multirow{2}{*}{\raisebox{-0.5pt}{\textbf{mIoU}}} \\
        & &\textbf{Dice} &\textbf{IoU} &\textbf{Dice} &\textbf{IoU} &\textbf{Dice} &\textbf{IoU} &\textbf{Dice} &\textbf{IoU} &\textbf{Dice} &\textbf{IoU} &\textbf{Dice} &\textbf{IoU} &\textbf{Dice} &\textbf{IoU} &\textbf{Dice} &\textbf{IoU} & & \\
\hline
    - &- &67.76 &56.36 &95.57 &91.87 &26.78 &20.06 &38.92 &29.58 &36.92 &27.54 &72.34 &72.34 	&79.88 &69.96 &63.19 &50.42 &60.17 &52.27 
\\
    - &$\checkmark$ &72.47 &62.22 &95.35 &91.70 &17.75 &14.56 &41.41 &33.47 &41.26 &31.81 &73.54 	&63.44 &80.80 &79.56 &69.72 &54.41 &61.54 &53.90 
\\
    $\checkmark$ &- &67.67 &56.90 &95.18 &91.26 &83.52 &73.45 &88.56 &80.08 &77.85 &64.31 &73.53 	&62.32 &80.80 &70.80 &61.95 &49.39 &78.63 &68.56 
\\
    $\checkmark$ &$\checkmark$ &73.71 &63.76 &95.65 &92.17 &86.63 &77.58 &91.11 &83.94 &81.31 &69.05 &73.49 &63.83 &81.21 &71.14 &65.96 &54.22 &81.13 &71.96 
\\
\hline

    \end{tabular}%
    }
  \caption{Ablation study across eight ultrasound image segmentation tasks under the 1/16 data partition. PuD denotes prompting-upon-decoding.}
  \label{tab:ablation116}%
\end{table*}%

\begin{table*}[h!]
  \centering
  \large
  \def\arraystretch{1.3}
  
    \resizebox{1\textwidth}{!}{
    \begin{tabular}{>{\centering}m{1.2cm} >{\centering}m{1.2cm}|cc|cc|cc|cc|cc|cc|cc|cc|c|c}
\hline
    \multirow{2}{*}{\raisebox{-0.5pt}{\textbf{PuD}}} &\multirow{2}{*}{\raisebox{-0.5pt}{\textbf{UPLC}}} &\multicolumn{2}{c|}{\raisebox{-0.5pt}{\textbf{BC}}} &\multicolumn{2}{c|}{\raisebox{-0.5pt}{\textbf{FH}}} &\multicolumn{2}{c|}{\raisebox{-0.5pt}{\textbf{LA}}} & \multicolumn{2}{c|}{\raisebox{-0.5pt}{\textbf{LV}}} & \multicolumn{2}{c|}{\raisebox{-0.5pt}{\textbf{MD}}} & \multicolumn{2}{c|}{\raisebox{-0.5pt}{\textbf{OT}}} & \multicolumn{2}{c|}{\raisebox{-0.5pt}{\textbf{TG}}} & \multicolumn{2}{c|}{\raisebox{-0.5pt}{\textbf{TN}}} & \multirow{2}{*}{\raisebox{-0.5pt}{\textbf{mDice}}} & \multirow{2}{*}{\raisebox{-0.5pt}{\textbf{mIoU}}} \\
    & &\textbf{Dice} &\textbf{IoU} &\textbf{Dice} &\textbf{IoU} &\textbf{Dice} &\textbf{IoU} &\textbf{Dice} &\textbf{IoU} &\textbf{Dice} &\textbf{IoU} &\textbf{Dice} &\textbf{IoU} &\textbf{Dice} &\textbf{IoU} &\textbf{Dice} &\textbf{IoU} & & \\
\hline
    - &- &71.86 &61.65 &95.55 &91.96 &21.13 &15.33 &63.78 &53.18 &26.71 &18.95 &74.25 &63.60 	&84.04 &74.44 &65.83 &53.91 &62.89 &54.13 
 
\\
    - &$\checkmark$ &70.30 &59.79 &94.10 &89.61 &16.35 &10.51 &68.97 &54.24 &40.30 &26.78 &74.93 	&64.75 &76.56 &65.98 &64.01 &52.17 &63.19 &52.98 
 
\\
    $\checkmark$ &- &68.97 &58.23 &96.35 &93.32 &87.68 &78.98 &91.78 &85.10 &83.35 &71.94 &74.38 	&64.13 &80.54 &72.19 &66.57 &53.97 &81.20 &72.23 

\\
    $\checkmark$ &$\checkmark$ &75.03 &65.29 &96.25 &93.20 &87.16 &78.85 &92.29 &85.90 &84.13 &73.08 &77.05 &66.91 &83.61 &75.59 &71.30 &59.83 &83.35 &74.83 
 
\\
\hline

    \end{tabular}%
    }
  \caption{Ablation study across eight ultrasound image segmentation tasks under the 1/8 data partition.}
  \label{tab:ablation18}%
\end{table*}%

\begin{table*}[h!]
  \centering
  \large
  \def\arraystretch{1.3}
  
    \resizebox{1\textwidth}{!}{
    \begin{tabular}{>{\centering}m{1.2cm} >{\centering}m{1.2cm}|cc|cc|cc|cc|cc|cc|cc|cc|c|c}
\hline
    \multirow{2}{*}{\raisebox{-0.5pt}{\textbf{PuD}}} &\multirow{2}{*}{\raisebox{-0.5pt}{\textbf{UPLC}}} &\multicolumn{2}{c|}{\raisebox{-0.5pt}{\textbf{BC}}} &\multicolumn{2}{c|}{\raisebox{-0.5pt}{\textbf{FH}}} &\multicolumn{2}{c|}{\raisebox{-0.5pt}{\textbf{LA}}} & \multicolumn{2}{c|}{\raisebox{-0.5pt}{\textbf{LV}}} & \multicolumn{2}{c|}{\raisebox{-0.5pt}{\textbf{MD}}} & \multicolumn{2}{c|}{\raisebox{-0.5pt}{\textbf{OT}}} & \multicolumn{2}{c|}{\raisebox{-0.5pt}{\textbf{TG}}} & \multicolumn{2}{c|}{\raisebox{-0.5pt}{\textbf{TN}}} & \multirow{2}{*}{\raisebox{-0.5pt}{\textbf{mDice}}} & \multirow{2}{*}{\raisebox{-0.5pt}{\textbf{mIoU}}} \\
    & &\textbf{Dice} &\textbf{IoU} &\textbf{Dice} &\textbf{IoU} &\textbf{Dice} &\textbf{IoU} &\textbf{Dice} &\textbf{IoU} &\textbf{Dice} &\textbf{IoU} &\textbf{Dice} &\textbf{IoU} &\textbf{Dice} &\textbf{IoU} &\textbf{Dice} &\textbf{IoU} & & \\
\hline
    - &- &70.09 &58.36 &94.66 &90.25 &20.02 &12.07 &70.18 &54.81 &39.95 &25.98 &74.35 &62.87 &	78.92 &67.88 &64.18 &51.57 &64.04 &52.97

\\
    - &$\checkmark$ &70.71 &59.18 &94.93 &90.87 &31.71 &19.59 &66.32 &50.35 &36.49 &23.37 &74.75 	&63.96 &75.57 &63.50 &66.91 &54.73 &64.67 &53.19

\\
    $\checkmark$ &- &75.15 &65.33 &96.37 &93.36 &87.73 &79.41 &92.65 &86.46 &84.28 &73.29 &78.31 	&68.46 &87.19 &79.39 &69.02 &57.38 &83.84 &75.39

\\
    $\checkmark$ &$\checkmark$ &77.18 &67.91 &96.56 &93.64 &88.72 &80.83 &93.00 &87.07 &85.24 	&74.70 &78.31 &68.86 &89.52 &82.14 &75.44 &64.50 &85.50 &77.46

\\
\hline

    \end{tabular}%
    }
  \caption{Ablation study across eight ultrasound image segmentation tasks under the 1/4 data partition.}
  \label{tab:ablation14}%
\end{table*}%

\subsection{Comparison with State-of-the-Art Methods}
We compare ProPL against four categories of methods:
\begin{itemize}
\item Single-task supervised: U-Net \cite{ref_unet}, PSPNet \cite{ref_pspnet}, Swin-Unet \cite{ref_swinunet}, Mamba-UNet \cite{ref_mambaunet}, DeepLabv3+ \cite{ref_deeplabv3+}, SegFormer \cite{ref_segformer}, Swin UNETR \cite{ref_swinunetr}, HiFormer \cite{ref_hiformer}, H2Former \cite{ref_h2former}, and UNet++ \cite{ref_unet++}
\item Single-task semi-supervised: FixMatch \cite{ref_fixmatch}, UniMatch \cite{ref_unimatch}, U$^2$PL \cite{ref_U2PL}, AugSeg~\cite{ref_augseg}, DDFP~\cite{ref_ddfp}, CSC-PA~\cite{ref_cscpa}, and ST$++$ \cite{ref_ST}
\item Universal supervised: DoDNet \cite{ref_dodnet} and CLIP-UM \cite{ref_clip}
\item Universal semi-supervised: Univ-full \cite{ref_usss}
\end{itemize}
\par
Note that \citet{ref_versemi} present pioneering work for universal semi-supervised medical image segmentation on CT/MRI images and achieve impressive results. However, in our experiments, it performs mediocrely on ultrasound images, so we did not include it. Table \ref{tab:comparison116} presents results across all eight segmentation tasks under the 1/16 partition protocol. ProPL demonstrates substantial improvements over all competitors, surpassing the best single-task supervised model by 5.18\% in mean Dice and 5.87\% in mIoU, the top single-task semi-supervised model by 0.97\%, the leading universal supervised model by 17.36\% and 20.59\%, and Univ-full by 33.74\% and 35.98\%. We conduct additional comparison experiments under 1/8 and 1/4 data partitions (Tables \ref{tab:comparison18} and \ref{tab:comparison14}). Under 1/8 data partitioning, our model outperforms the second-best model by 2.2\% in mean Dice, while under 1/4 data partitioning, our model achieves improvements of 1.32\% in mean Dice and 1.82\% in mean IoU compared to the second-best model. 
\par
Furthermore, our model also delivers strong performance on other metrics. For example, in terms of HD95 (lower is better), our method attains 20.59, outperforming the second-best result by 1.06 under the 1/8 setting.

\par
Figure \ref{fig:vision} shows qualitative comparisons between our model, Mamba-UNet, U$^2$PL, CLIP-UM, and Univ-full.

\subsection{Discussion}

\noindent
\textbf{Task-Specific Prompting}
Removing task prompts leads to performance drops of 19.59\% and 18.06\% in mean Dice and mIoU under 1/16 data partition, respectively (see Table \ref{tab:ablation116}). We can also observe similar degradation under 1/8 and 1/4 data partitions (see Tables \ref{tab:ablation18} and \ref{tab:ablation14}).

\noindent
\textbf{Prompting-upon-Decoding}
Our approach outperforms existing prompting methods implemented in CLIP-UM~\cite{ref_clip} and DoDNet~\cite{ref_dodnet} across all eight tasks (Figure \ref{fig:PUD}), demonstrating gains of 3.05\% and 1.16\% in mean Dice, respectively.

\noindent
\textbf{UPLC Impact}
The absence of pseudo-label calibration reduces performance by 2.5\% and 3.4\% in average Dice and IoU under 1/16 data partition (cf. Table \ref{tab:ablation116}). We can also see consistent performance degradation under 1/8 and 1/4 data partitions (cf. Tables \ref{tab:ablation18} and \ref{tab:ablation14}).

\noindent
\textbf{Stochastic Perturbations in UPLC}
We investigate the impact of varying stochastic perturbation counts in UPLC. As shown in Figure \ref{fig:passtime}, optimal performance is achieved with two perturbations under 1/4 and 1/8 partition protocols (85.61\% and 83.43\% respectively) and three perturbations under the 1/16 protocol (81.13\%). GPU memory constraints prevented testing beyond four perturbations. Based on these findings, we adopt two stochastic perturbations in our implementation.
\par
Increasing $N$ by 1 incurs an additional 2.43 GB of memory and 27.6 s per epoch, which remains practically acceptable.
\par
We also conduct comparative experiments on random perturbation strategies, evaluating Dropout against Gaussian noise. The results show that Gaussian noise performs worse, with a 2.58 mIoU drop.

\noindent
\textbf{Memory Footprint}
We compare ProPL's memory efficiency against representative methods including U-Net, Mamba-UNet, U$^2$PL, DoDNet, CLIP-UM, and Univ-full. As shown in Figure \ref{fig:memory}, ProPL achieves the highest Dice while maintaining a modest memory footprint of 712 MB, demonstrating optimal performance-efficiency trade-off.

\noindent
\textbf{Prompting Trade-Off}
We use minimal prompts at negligible generation cost. Removing them causes collapse (Tables \ref{tab:ablation116}, \ref{tab:ablation18}, \ref{tab:ablation14}), making extra-computation metrics meaningless. Compared to one-hot prompting, ours adds only 18 s/epoch but improves results---an acceptable trade-off.

\noindent
\textbf{Generalization}
Public ultrasound datasets are scarce. On an in-house external set (700 images), the model achieves 73.33 mIoU---comparable to internal results---showing good generalization.

\section{Conclusion}
In this paper, we address the challenging problem of universal semi-supervised ultrasound image segmentation by introducing ProPL, a framework that leverages prompt-guided pseudo-labeling to effectively segment multiple anatomical structures while utilizing both labeled and unlabeled data. Extensive experiments on our newly compiled ultrasound dataset demonstrate that ProPL consistently outperforms state-of-the-art methods.

\section{Acknowledgments}
This work was supported in part by the National Key Research and Development Program of China (Grant 2022ZD0160604), the Hainan Province ``Nanhai New Star'' Technology Innovation Talent Platform Project (Grant NHXXRCXM202361), the Sanya Science and Technology Special Fund (Grant 2024KJFX034), the NSFC (Grant 62571381), the General Program of the Hubei Natural Science Foundation (Grant 2025AFB615), and the Key Research and Development and Achievement Transformation Program of the Inner Mongolia Autonomous Region (Grant 2025YFHH0078).

\bibliography{aaai2026}

\end{document}